\documentclass[10pt,twocolumn,letterpaper]{article}

\usepackage[pagenumbers]{cvpr}      

\usepackage{graphicx}
\usepackage{amsmath}
\usepackage{amssymb}
\usepackage{booktabs}
\usepackage{pifont}
\usepackage{multirow}
\usepackage{tabu}
\usepackage{color}
\usepackage{marvosym}

\usepackage[pagebackref,breaklinks,colorlinks]{hyperref}

\usepackage[capitalize]{cleveref}
\crefname{section}{Sec.}{Secs.}
\Crefname{section}{Section}{Sections}
\Crefname{table}{Table}{Tables}
\crefname{table}{Tab.}{Tabs.}

\newcommand{\cmark}{\ding{51}}
\newcommand{\xmark}{\ding{55}}
\def\cvprPaperID{6947} 
\def\confName{CVPR}
\def\confYear{2023}

\makeatletter
\def\blfootnote{\xdef\@thefnmark{}\@footnotetext}
\makeatother
\begin{document}

\title{HSTFormer: Hierarchical Spatial-Temporal Transformers \\ for 3D Human Pose Estimation}

\author{\fontsize{11.3pt}{\baselineskip}\selectfont Xiaoye Qian$^1$$^\ast$, Youbao Tang$^2$$^\ast$\textsuperscript{\Letter}, Ning Zhang$^2$, Mei Han$^2$, Jing Xiao$^3$, Ming-Chun Huang$^4$, Ruei-Sung Lin$^2$\\
$^1$Case Western Reserve University, $^2$PAII Inc., $^3$Ping An Technology, $^4$Duke Kunshan University\\
}
\maketitle

\blfootnote{\noindent $^{*}$The first two authors have equal contribution. This work was done during Xiaoye's intership in PAII Inc..
\textsuperscript{\Letter}Corresponding author.}

\begin{abstract}
   Transformer-based approaches have been successfully proposed for 3D human pose estimation (HPE) from 2D pose sequence and achieved state-of-the-art (SOTA) performance. However, current SOTAs have difficulties in modeling spatial-temporal correlations of joints at different levels simultaneously. This is due to the poses' spatial-temporal complexity. Poses move at various speeds temporarily with various joints and body-parts movement spatially. Hence, a cookie-cutter transformer is non-adaptable and can hardly meet the ``in-the-wild" requirement. To mitigate this issue, we propose \textbf{H}ierarchical \textbf{S}patial-\textbf{T}emporal trans\textbf{Former}s (HSTFormer) to capture multi-level joints' spatial-temporal correlations from local to global gradually for accurate 3D HPE. HSTFormer consists of four transformer encoders (TEs) and a fusion module. To the best of our knowledge, HSTFormer is the first to study hierarchical TEs with multi-level fusion. Extensive experiments on three datasets (\textit{i.e.}, Human3.6M, MPI-INF-3DHP, and HumanEva) demonstrate that HSTFormer achieves competitive and consistent performance on benchmarks with various scales and difficulties. Specifically, it surpasses recent SOTAs on the challenging MPI-INF-3DHP dataset and small-scale HumanEva dataset,  with a highly generalized systematic approach. The code is available at: {\tt\small \url{https://github.com/qianxiaoye825/HSTFormer}}.
\end{abstract}

\section{Introduction}
\label{sec:intro}

\begin{figure}[ht]
	\centering
	\includegraphics[width=0.48\textwidth]{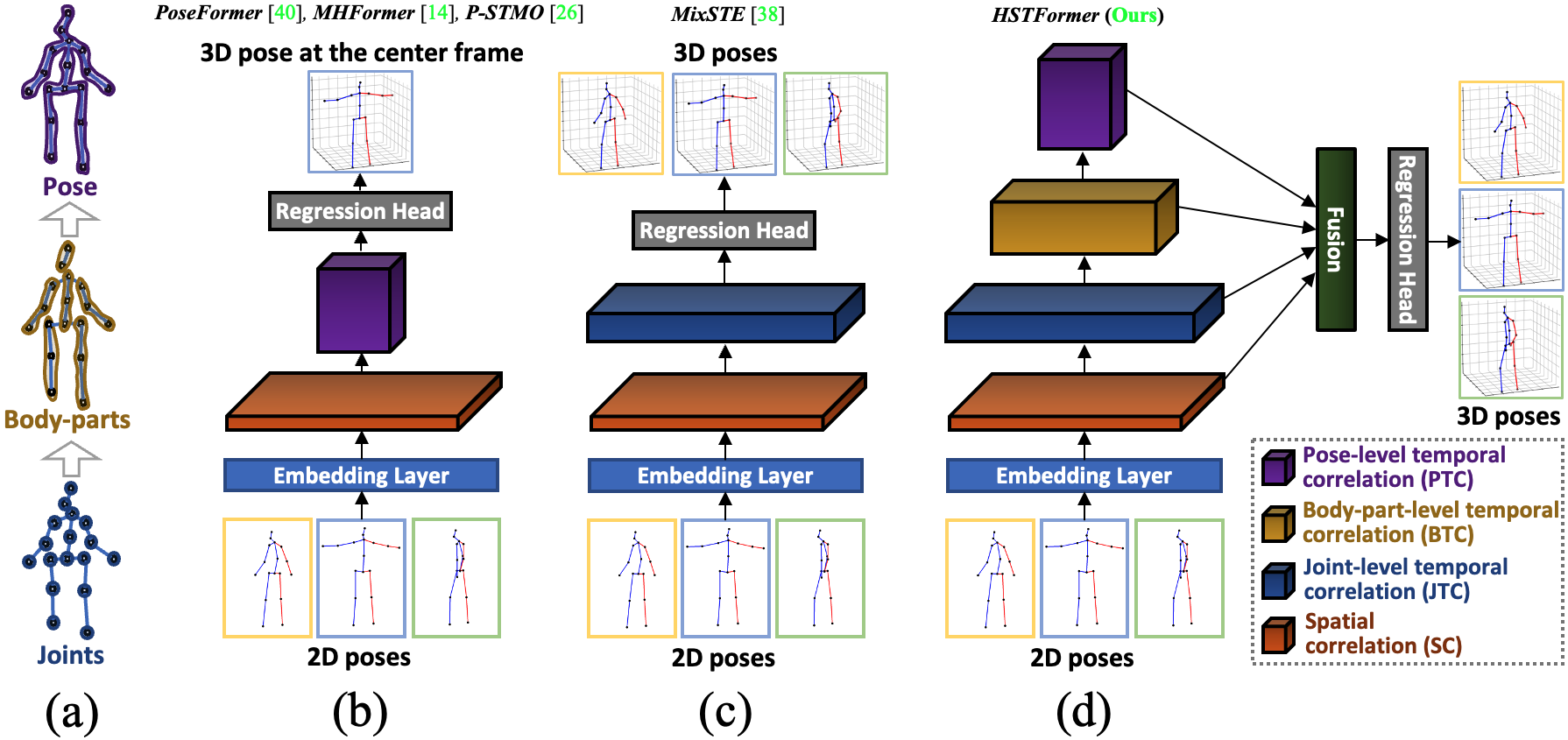}
    \caption{(a) The hierarchical bottom-up structure of pose and (b)-(d) the basic architectures of different transformer-based 3D HPE methods. Previous approaches focus on specific correlations without a systematic and structural analysis. For example, (b) PoseFormer \cite{poseformer}, MHformer \cite{li2022mhformer}, and P-STMO \cite{P-STMO} only consider \textcolor[RGB]{163,95,51}{SC} and \textcolor[RGB]{90,29,134}{PTC}. (c) MixSTE \cite{zhang2022mixste} only considers \textcolor[RGB]{163,95,51}{SC} and \textcolor[RGB]{69,94,135}{JTC}. (d) Ours based on hierarchical bottom-up transformer encoders to learn multi-level spatial-temporal correlations of joints systematically, including all of \textcolor[RGB]{163,95,51}{SC}, \textcolor[RGB]{69,94,135}{JTC}, \textcolor[RGB]{167,122,2}{BTC}, and \textcolor[RGB]{90,29,134}{PTC}, and adaptively integrate these multi-level correlations using a fusion module. The figure is best viewed in color.}
	\label{fig.hierarchical}
\end{figure}

3D human pose estimation (HPE) aims to predict the coordinates of human body joints in 3D space from images and videos. Hence, it plays important roles in numerous applications, such as human activity recognition \cite{liu2018recognizing}, healthcare \cite{huang2018video}, augmented reality (AR) \cite{guler2018densepose}, virtual reality (VR) \cite{mehta2017vnect} and human-robot interaction \cite{munea2020progress}. Direct inferring 3D pose from a single image is a challenging ill-posed problem because part of the 3D information is lost in 3D to 2D projection. An alternative approach for 3D HPE is from video, given that human pose changes smoothly and structurally through time. With the advances in deep learning, estimating 2D human poses has become a standard technique. By first converting the human image sequence to 2D pose sequences, 3D HPE from the video can be reduced to the problem of inferring a 3D pose sequence from its 2D counterpart. 

To solve this problem, earlier approaches\cite{cai2019exploiting,dabral2018learning,videopose3d,wang2020motion} use dilated temporal convolution or graph convolution networks to exploit spatial and temporal representations of 2D pose sequences. However, they generally rely on temporal dilation techniques or a predefined adjacency matrix to model temporal relationships. Consequently, the temporal connectivity inside these models is still limited. 

Recently, transformer \cite{vaswani2017attention} with self-attention mechanism is shown capable of capturing long-range relationships of the input sequences, and has been successfully applied to many challenging computer vision tasks \cite{detr,dosovitskiy2020image,yang2021transpose,poseformer}. The 2D-to-3D pose sequence estimation fits naturally to the transformer. As a matter of fact, most models that achieve current SOTA results are based on transformers \cite{poseformer,li2022mhformer, P-STMO,zhang2022mixste}. Among these models, PoseFormer \cite{poseformer} is a seminal work followed by MHFormer \cite{li2022mhformer} and P-STMO \cite{P-STMO}. This group considers all joints of each frame as a whole to learn the global pose-level temporal correlation across all frames. Their basic architecture is shown in Figure \ref{fig.hierarchical}(b).  
As a separate group itself, MixSTE \cite{zhang2022mixste}'s basic architecture is shown in Figure \ref{fig.hierarchical}(c). It decomposes a pose sequence into multiple joint sequences and uses stacks of spatial-temporal transformers to infer 3D poses by learning the local joint-level temporal correlation of each joint separately. These demonstrate that the temporal correlations of joints can be further dug to improve 3D HPE performance. 

However, as Figure \ref{fig.hierarchical} illustrates, aforementioned transformer-based SOTA approaches focus on spatial nuances and hence are lack of a systematic information propagation from local joints to global poses temporarily. On the other hand, we are inspired from the residual neural network and vision transformers developments, such as the Feature Pyramid Network (FPN) \cite{lin2017feature} improves the ResNet\cite{he2016deep}, as well as Pyramid Vision Transformer (PVT) \cite{wang2021pyramid} advances the Vision Transformer (ViT) \cite{dosovitskiy2020image}. We found that both the FPN and the PVT in an hierarchical and systematic pyramid structure, improve their predecessors with better information propagation and consequently better task results.  

Motivated by the literature, we propose a novel framework based on spatial-temporal transformers that leverages the hierarchical bottom-up structure of a pose shown in Figure \ref{fig.hierarchical}(a). A pose is a composition of a set of articulated body parts, and each body part is a collection of corresponding joints. Analyzing a pose sequence is performed bottom-up in this hierarchy. First, joint sequences are independently passed through a temporal transformer. Then, the outputs of joint sequences are aggregated to become body part sequences and pass through temporal transformers for different body parts. These outputs are aggregated again to become the pose sequence as input to a temporal transformer for the pose. Finally all the outputs from the joints, parts, and pose transformer are fused to adaptively integrate the multi-level complementary information for 3D pose estimation. This hierarchically bottom-up approach enables us to exploit the structural nature of human poses and extract valuable information locally and globally. Figure \ref{fig.hierarchical}(d) illustrates the systematic hierarchical spatial-temporal transformers. Specifically, bottom-up structural transformer encoders are organized to learn different correlations of joints from local to global as follows: (i) the Spatial Correlation (SC) of an individual single frame; (ii) the Joint-level Temporal Correlation (JTC) cross frames; (iii) the Body-part-level  Temporal Correlation (BTC), which groups joints correlation across frames; and (iv) the Pose-level Temporal Correlation (PTC) of all joints across frames. Compared to the previous approaches illustrated in Figure \ref{fig.hierarchical}(b) \cite{poseformer,li2022mhformer,P-STMO}, and Figure \ref{fig.hierarchical}(c) \cite{zhang2022mixste}, which are less structurally and sporadically use transformer encoders, HSTFormer is able to model the joints' spatial-temporal correlations more comprehensively and systematically. Our main contribution can be summarized as follows: 
\begin{itemize}
 
\item A novel transformer-based framework, hierarchical spatial-temporal transformers (HSTFormer), is proposed to model multiple levels of joints' spatial-temporal correlations structurally in a bottom-up fashion. It is able to propagate the joints' movement information smoothly and effectively. Such a framework is capable of accurately estimating 3D human poses in both simple and complex scenes following the 2D-to-3D lifting pipeline.
\item A body-part temporal transformer encoder is proposed to tackle the grouped joints correlation across frames. To the best of our knowledge, this is the first study in 3D pose estimation that uses transformer to focus on grouped joints correlation cross-temporal realm. The proposed design solely reduces MPJPE error with a significant amount of $4.7$ \textit{mm} ($11.8\%$).
\item Extensive experiments on Human3.6M, MPI-INF-3DHP, and HumanEva datasets are conducted to demonstrate the superior performance and high generalization ability of the proposed HSTFormer for 3D human pose estimation. Specifically, it outperforms the existing SOTA approaches by a remarkable margin on the challenging MPI-INF-3DHP dataset, decreasing the MPJPE by 24.6\% (from 54.9 \textit{mm} to 41.4 \textit{mm}).
\end{itemize}

\begin{figure*}[!t]
	\centering
	\includegraphics[width=0.99\textwidth]{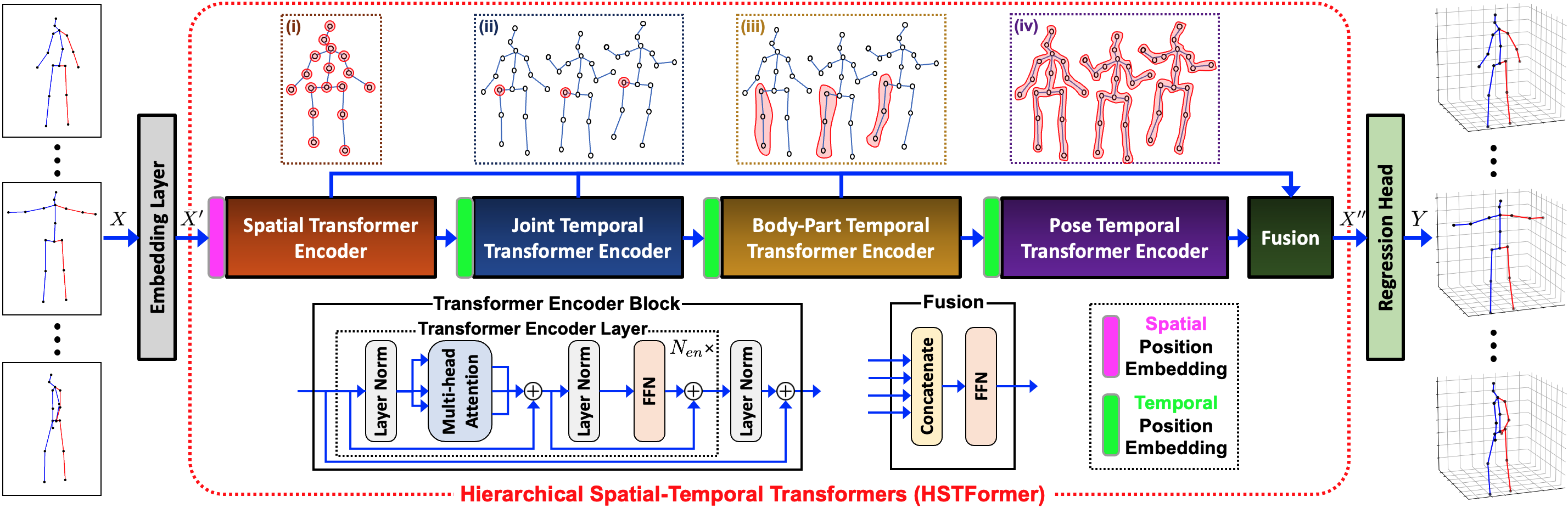}
	\caption{ System overview. 
    On the top of HSTFormer, the red regions represent the attended tokens for self-attention computation. For example, all joints within each frame are considered as tokens for the spatial transformer encoder, as shown in (i). The joints in a body part are concatenated to form a token for the body-part temporal transformer encoder, as shown in (iii).
}
	\label{fig.flowchart}
\end{figure*}

\section{Related Work}
Existing solutions to this problem can be divided into two categories: direct estimation approaches and 2D-to-3D lifting approaches. Direct estimation approaches \cite{pavlakos2017coarse,jin2022single} designed end-to-end frameworks to estimate the joints' 3D coordinates directly from images or videos without intermediate 2D pose representations, which are straightforward ways but remain challenges due to the lack of sufficient 3D in-the-wild data. 2D-to-3D lifting approaches \cite{zeng2020srnet,videopose3d,liu2020attention,poseformer,li2022mhformer,zhang2022mixste} have two stages which first detect 2D pose joints and then lift them to 3D.
Many existing works are based on the 2D-to-3D lifting pipeline and currently outperform the direct estimation approaches. Meanwhile, their generalizability is improved because of ignoring the persons' appearances when lifting. However, the 2D-to-3D lifting approaches also remain challenges due to occlusion and depth ambiguity in 2D poses. Especially, it is difficult to predict a 3D pose accurately only based on the spatial information of a 2D pose from a single frame \cite{martinez2017simple}.

To alleviate such an issue and to improve accuracy and robustness, many approaches \cite{cai2019exploiting,dabral2018learning,videopose3d,wang2020motion,poseformer,li2022mhformer,zhang2022mixste} have integrated temporal information from 2D pose sequences to explore the joints' spatial and temporal correlations. For example, Hossain and Little \cite{hossain2018exploiting} applied the Long Short-Term Memory (LSTM) to explore the temporal information. Pavllo et al. \cite{videopose3d} proposed a dilated temporal convolution network to capture global contextual information and estimate 3D pose from consecutive 2D sequences. Cai et al. \cite{cai2019exploiting} utilize graph convolution networks to exploit spatial and temporal graph representations of human skeletons for 3D pose estimation. However, convolutional neural network-based approaches typically rely on temporal dilation techniques and graph convolution network-based approaches depend on a predefined adjacency matrix to model spatial-temporal relationships. Eventually, the spatial-temporal connectivity explored by them is still limited.

Based on the mechanism of the transformer, the correlations through long sequence data can be extracted, which makes it possible to explore the temporal representations across the video frames for 3D HPE. 
Spatial and temporal transformers are gradually applied for 3D HPE \cite{poseformer,li2022mhformer,zhang2022mixste,P-STMO}.
PoseFormer \cite{poseformer} explored the spatial joint correlations from each frame and the temporal joint across frames by using  a spatial-temporal transformer structure, which is the first purely transformer-based approach for 3D HPE. MHFormer \cite{li2022mhformer} was further proposed to learn the spatial-temporal information by presenting a multi-hypothesis transformer. MixSTE \cite{zhang2022mixste} developed a mixed spatial-temporal transformer to extract the all joints correlation across frames by applying the spatial transformer encoder and temporal transformer encoder alternatively. P-STMO \cite{P-STMO} presented a spatial temporal many-to-one model which is pre-trained by using self-supervised learning for performance improvement. Our designs are inspired by the aforementioned works while making the model have the capability to learn the different levels of spatial-temporal information. Unlike previous works, we design two more modules to extract the spatial-temporal correlation of a group of joints across frames and adaptively integrate the learned multi-level spatial-temporal information of joints for accurate 3D HPE.

\section{Hierarchical Spatial-Temporal Transformers }
Figure \ref{fig.flowchart} shows the overview of the proposed 3D human pose estimation method which follows the popular 2D-to-3D lifting pipeline. An embedding layer first converts the input 2D pose sequence to high dimensional representations. 
Let $X$ be the input 2D pose sequence $X \in \mathbb{R}^{T \times J \times 2}$
with $T$ frames and $J$ joints per frame. $X$ obtained by an off-the-shelf 2D pose detector from a video.
The embedding layer maps $X$ to high-dimensional representations $X^{\prime} \in \mathbb{R}^{T \times J \times D}$ by: 
\begin{equation}\label{eqn:x'}
    X^{\prime} = LN\left(XW_L\right)
\end{equation}
where $W_L \in \mathbb{R}^{2 \times D}$ is a linear projection and $LN(*)$ means layer normalization. In this work, we set $D=32$ to trade off the computational efficiency and performance as processed in \cite{poseformer}.

Next, a cascade of different types of transformer encoders process these features in spatial and temporal domain. 
All these transformer encoders are based on the typical multi-head attention transformer encoder block (TE).
Spatial transformer encoder (STE) processes the joint features in each frame independently. 
Joint temporal transformer encoder (JTTE) works on the temporal feature sequence of each joint separately.  
Body-part temporal transformer encoder (BTTE) takes grouped joint temporal feature sequences that corresponding to a particular body part as input. 
Pose temporal transformer encoder (PTTE) groups all joint temporal feature sequences into a temporal sequence as its input.
These encoders have multiple layers, and we use $N_{en}$ to denote the number of layers. In summary, the propsoed STE, JTTE, BTTE and PTTE in Figure \ref{fig.flowchart} correspond to the SC, JTC, BTC and PTC in Figure \ref{fig.hierarchical}(d), respectively.

\begin{figure}[t]
	\centering
	\includegraphics[width=0.48\textwidth]{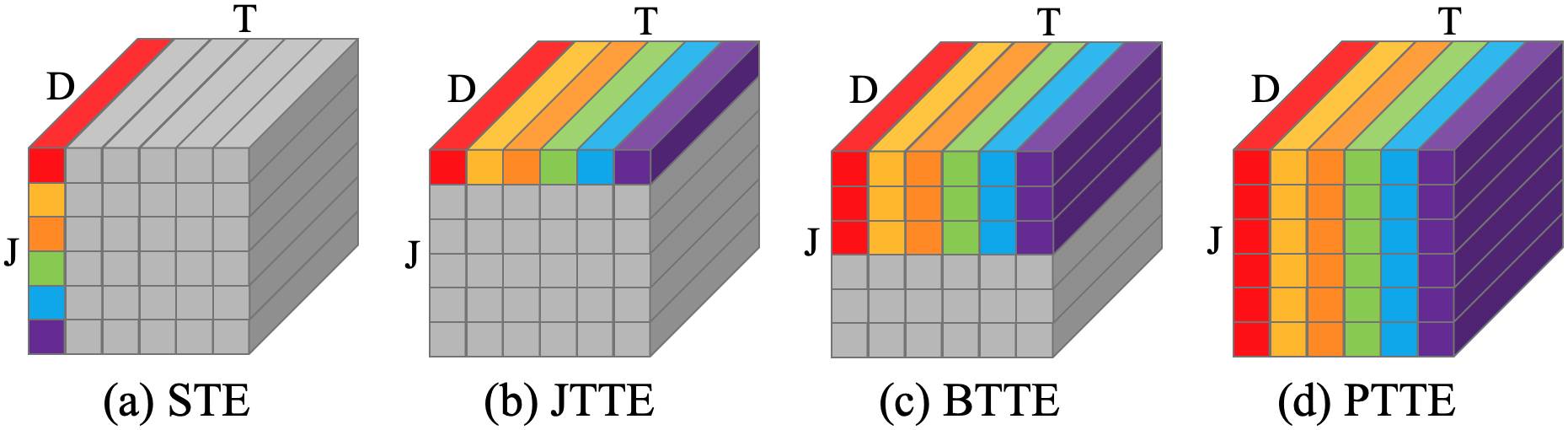}
	\caption{Tensor slicing performed by the four transformer encoders in the proposed HSTFormer. Different colors indicate different individual tokens attended for self-attention computation.}
	\label{fig.tensor}
\end{figure}

The following sections explain each of the aforementioned encoders. 
If we represent input features 
 as a tensor with shape $T \times J \times D$,  each encoder slices this tensor 
along different axes and with different block sizes as a sequence of input. Before diving into the details, Figure \ref{fig.tensor} provides an overall illustration on how the slicing is performed at each transformer encoder.

\subsection{Spatial Transformer Encoder}
\label{ste}
The spatial transformer encoder (STE) is designed to learn the spatial correlation among the joints within each frame. 
It treats 2D joint representations as the input sequence to STE. 
Let $x^i \in \mathbb{R}^{J \times D}$ be the features of all joints in the $i^{th}$ frame sliced from $X^{\prime} \in \mathbb{R}^{T \times J \times D}$.
A learnable positional encoding $E_{sp} \in \mathbb{R}^{J \times D}$ is added to $x^i$ before it is fed into STE.
After adding the matrix $E_{sp}$, $x^i$ is fed into STE to capture the spatial-level correlation among all joints in the $i^{th}$ frame using the self-attention mechanism. 
The output of STE will be $ Z_{ste} \in \mathbb{R}^{T \times J \times D} $ after processing all frames as follows:
\begin{equation}
    Z_{ste} = R(Concat(STE(x^1),...,STE(x^T)))
\end{equation}
where $R(*)$ is a rearrange operation.

\subsection{Joint Temporal Transformer Encoder}
\label{jtte}
As mentioned in \cite{zhang2022mixste}, the motion trajectories of the different body joints vary for different frames and should be learned separately. In this work, we also design a joint temporal transformer encoder (JTTE) to learn the joint-level temporal correlation by considering temporal motion trajectory information of each joint across all input frames independently.
Let $z_{ste}^j \in \mathbb{R}^{T \times D}$ be the the $j^{th}$ joint sequence from $Z_{ste}$
A learnable temporal positional encoding $E_{tp} \in \mathbb{R}^{T \times D}$ is added to $z_{ste}^j$ before it enters JTTE. 
JTTE applies the self-attention mechanism to enhance $z_{ste}^j$, where $i=1, \ldots, J$.
The output of JTTE from all joints will be strack together and rearrange to become $Z_{jtte} \in \mathbb{R}^{T \times J \times D}$.

\subsection{Body-Part Temporal Transformer Encoder}
\label{btte}
$Z_{jtte}$ is the joint feature sequence enhanced by considering the spatial correlation of different joints in each frame and the temporal correlation of each joint sequence separately.
As a fact, a pose is a composition of a set of articulated body parts and each body part is a collection of corresponding joints which are most relevant to each other. Generally, no matter simple or complex, the movements can be performed by the combined local motions of different body parts. Therefore, the global pose motion can be well captured by modeling the local body-part motions accurately and separately.
Here we divide $Z_{jtte}$ into six sequences according to different body parts: Head: \{Nose, Head\}, Torso: \{Hip, Spine, Thorax\}, Left Hand: \{LShoulder, LElbow, LWrist\}, Right Hand: \{RShoulder, RElbow, RWrist\}, Left Leg: \{LHip, LKnee, LFoot\}, and Right Leg: \{RHip, RKnee, RFoot\}.
We then apply six body-part temporal transformer encoders (BTTE), one for each specific body part, to these six sequences respectively. 
These BTTEs are designed to capture the temporal correlation of each body part sequence. 

BTTE takes $ Z_{jtte} \in \mathbb{R}^{T \times J \times D} $ as input. It is split into $6$ sequences, $\left\{Z_{jtte}^k \in \mathbb{R}^{T \times J^{k}D}|k=1,...,6\right\}$, where $J^k$ is the number of joints in the $k^{th}$ group.
That is, the representations of joints from the same body parts are concatenated to form new frame-level features. As done in JTTE, a learnable temporal positional encoding $E_{tp} \in \mathbb{R}^{T \times D}$ is added to $Z_{jtte}^k$. 
The output of 6 BTTEs will be stack and rearrange to the orignal shape, $ Z_{btte} \in \mathbb{R}^{T \times J \times D} $.

\subsection{Pose Temporal Transformer Encoder}
\label{ptte}
$Z_{btte}$ captures the inter-joint spatial-temporal correlation up to the body-part level. Following the work \cite{poseformer,li2022mhformer,P-STMO}, we design a pose temporal transformer encoder (PTTE) to capture the global pose-level temporal correlation of all joints across frames. 
PTTE reshapes $ Z_{btte}$ as $ Z_{btte} \in \mathbb{R}^{T \times J D} $, which concatenates all joint features in a single frame as a new feature. The output of PTTE will be $ Z_{ptte} \in \mathbb{R}^{T \times J \times D} $ after rearranging.

\subsection{Fusion Module}
\label{fm}
Following the above four TEs is a fusion layer that collects the output from these encoders and creates the final features, which is sent to the regression head for 3D pose prediction. The output of these TEs capture different-level of granularity of the spatial and temporal information extracted from the input 2D pose sequence. To sufficiently utilize their complementary information, a fusion module is designed for adaptively integrating the outputs of all four encoders. It is simply implemented by a fully connected feed-forward network with a learnable weight $W_F \in \mathbb{R}^{4D \times D}$.The final enhanced high-dimensional representations $X^{\prime\prime} \in \mathbb{R}^{T \times J \times D}$ can be obtained using the following procedures:
\begin{equation}
    X^{\prime\prime} = Concat(Z_{ste},Z_{jtte},Z_{btte},Z_{ptte})W_F
\end{equation}

\subsection{Regression Head}
Our method performs sequence to sequence prediction, so the regression head is used to project the high-dimension output of the fusion module $X^{\prime\prime} \in \mathbb{R}^{T \times J \times D}$ to 3D coordinates of the pose sequence $Y \in \mathbb{R}^{T \times J \times 3}$ by applying the linear transformation layer. 

\subsection{Loss Function}
As done in \cite{poseformer}, the standard MPJPE (Mean Per Joint Position Error) loss is used to train our entire model in an end-to-end manner by minimizing the error between the predicted and ground truth pose sequence as:
\begin{equation}
    L = \sum_{i=1}^T\sum_{j=1}^J || Y_{j}^{i}-\widetilde{Y_{j}^{i}} ||_2
\end{equation}
where $Y_{j}^{i}$ and $\widetilde{Y_{j}^{i}}$ are the predicted and ground truth 3D coordinates of the $j^{th}$ joint in the $i^{th}$ frame, respectively.

\begin{table*}[htb]
  \normalsize
  \centering
  \caption
  {
    Quantitative comparison with the state-of-the-art methods on Human3.6M under Protocol 1 (MPJPE  (\textit{mm})), using detected 2D poses (top) and ground truth 2D poses (bottom) as inputs. 
    \textcolor{red}{Red}: best; \textcolor{blue}{Blue}: second best. $T$ is the input length (the number of frames). 
  } 
  \resizebox{\textwidth}{!}{
  \begin{tabular}{@{}l|c|ccccccccccccccc|c@{}}
  \toprule[1pt]
  Method (CPN) & & Dir. & Disc & Eat & Greet & Phone & Photo & Pose & Purch. & Sit & SitD. & Smoke & Wait & WalkD. & Walk & WalkT. & Avg.\\
  \midrule[0.5pt]
  GraphSH~\cite{xu2021graph} ($T$=1) & CVPR'21 &45.2 &49.9 &47.5 &50.9 &54.9 &66.1 &48.5 &46.3 &59.7 &71.5 &51.4 &48.6 &53.9 &39.9 &44.1 &51.9 \\
  GraFormer~\cite{zhao2022graformer} ($T$=1) & CVPR'22  &45.2 &50.8 &48.0 &50.0 &54.9 &65.0 &48.2 &47.1 &60.2 &70.0 &51.6 &48.7 &54.1 &39.7 & 43.1 &51.8 \\
  MGCN~\cite{zou2021modulated}($T$=1) & ICCV'21 &45.4 &49.2 &45.7 &49.4 &50.4 &58.2 &47.9 &46.0 &57.5 &63.0 &49.7 &46.6 &52.2 &38.9 &40.8 &49.4 \\
  ST-GCN~\cite{cai2019exploiting} ($T$=7) & ICCV'19  &44.6 &47.4 &45.6 &48.8 &50.8 &59.0 &47.2 &43.9&57.9 &61.9 &49.7 &46.6 &51.3 &37.1 &39.4 &48.8 \\
  VPose~\cite{videopose3d} ($T$=243) & CVPR'19  & 45.2 & 46.7 & 43.3 & 45.6 & 48.1 & 55.1 & 44.6 & 44.3 & 57.3 & 65.8 & 47.1 & 44.0 & 49.0 & 32.8 & 33.9 & 46.8 \\
  UGCN~\cite{wang2020motion} ($T$=96) & ECCV'20  &41.3 &43.9 &44.0 &42.2 &48.0 &57.1 &42.2 &43.2 &57.3 &61.3 &47.0 &43.5 &47.0 &32.6 &31.8 &45.6 \\
  Liu \emph{et al.}~\cite{liu2020attention} ($T$=243) & CVPR'20  &41.8 &44.8 &41.1 &44.9 &47.4 &54.1 &43.4 &42.2 &56.2 &63.6 &45.3 &43.5 &45.3 &31.3 &32.2 &45.1 \\
  PoseFormer~\cite{poseformer} ($T$=81) & ICCV'21  &{41.5} &44.8 &\textcolor{blue}{39.8} &42.5 &{46.5} &{51.6} &42.1 &{42.0} &{53.3} &{60.7} &45.5 &43.3 &46.1 &31.8 &32.2 &44.3 \\
  Anatomy3D~\cite{chen2021anatomy} ($T$=243) & TCSVT'21  &41.4 &{43.2} &{40.1} &42.9 &46.6 &{51.9} &{41.7} &42.3 &53.9 &{60.2} &45.4 &{41.7} &46.0 &31.5 &32.7 &{44.1} \\
  MHFormer~\cite{li2022mhformer} ($T$=351) & CVPR'22 & 39.2 & 43.1 & 40.1 & 40.9 & 44.9 & 51.2 & \textcolor{blue}{40.6} & 41.3 & \textcolor{blue}{53.5} & 60.3 & 43.7 & 41.1 & 43.8 & 29.8 & 30.6 & 43.0\\
  P-STMO~\cite{P-STMO} ($T$=243) & ECCV'22  &\textcolor{blue}{38.9}&42.7&40.4&41.1&45.6&\textcolor{red}{49.7}&40.9&\textcolor{blue}{39.9}&55.5&59.4&44.9&42.2&\textcolor{blue}{42.7}&\textcolor{blue}{29.4}&\textcolor{blue}{29.4}&42.8\\ 
  MixSTE~\cite{zhang2022mixste} ($T$=243) & CVPR'22  & \textcolor{red}{37.6}   & \textcolor{red}{40.9} & \textcolor{red}{37.3} & \textcolor{red}{39.7} & \textcolor{red}{42.3} & \textcolor{blue}{49.9}    & \textcolor{red}{40.1}    & \textcolor{red}{39.8} & \textcolor{red}{51.7}          & \textcolor{red}{55.0} & \textcolor{red}{42.1} & \textcolor{red}{39.8} & \textcolor{red}{41.0} & \textcolor{red}{27.9} & \textcolor{red}{27.9} & \textcolor{red}{40.9} \\
  \hline
  HSTFormer ($T$=81) & Ours  & 39.5 & \textcolor{blue}{42.0} & 39.9 & \textcolor{blue}{40.8} & \textcolor{blue} {44.4} & 50.9 &40.9 &41.3 & 54.7 & \textcolor{blue} {58.8} & \textcolor{blue}{43.6} & \textcolor{blue}{40.7} & 43.4 & 30.1 & 30.4 & \textcolor{blue}{42.7} \\
  \toprule[1pt]
  Method (GT) & & Dir. & Disc & Eat & Greet & Phone & Photo & Pose & Purch. & Sit & SitD. & Smoke & Wait & WalkD. & Walk & WalkT. & Avg.\\
  \midrule[0.5pt]
  VPose~\cite{videopose3d} ($T$=243) & CVPR'19  &35.2 &40.2 &32.7 &35.7 &38.2 &45.5 &40.6 &36.1 &48.8 &47.3 &37.8 &39.7 &38.7 &27.8 & 29.5 &37.8 \\
  GraFormer~\cite{zhao2022graformer} ($T$=1) & CVPR'22  &32.0 &38.0 &30.4 &34.4 &34.7 &43.3 &35.2 &31.4 &38.0 &46.2 &34.2 &35.7 &36.1 &27.4 & 30.6 &35.2 \\
  Liu \emph{et al.}~\cite{liu2020attention} ($T$=243) & CVPR'20  &34.5 &37.1 &33.6 &34.2 &32.9 &37.1 &39.6 &35.8 &40.7 &41.4 &33.0 &33.8 &33.0 &26.6 &26.9 &34.7 \\
  
  Ray3D \emph{et al.}~\cite{zhan2022ray3d} ($T$=9) & CVPR'22  &31.2 &35.7 &31.4 &33.6 &35.0 &37.5 &37.2 &30.9 &42.5 &41.3 &34.6 &36.5 &32.0 &27.7 &28.9 &34.4 \\
  SRNet~\cite{zeng2020srnet} ($T$=243) & ECCV'20  &34.8 &{32.1} &{28.5} &{30.7} &{31.4} &{36.9} &35.6 &{30.5} &{38.9} &40.5 &32.5 &{31.0} &29.9 &{22.5} &24.5 &32.0 \\
  PoseFormer~\cite{poseformer} ($T$=81) & ICCV'21  &{30.0} &{33.6} &{29.9} &31.0 &{30.2} &{33.3} &{34.8} &31.4 &{37.8} &{38.6} &{31.7} &{31.5} &{29.0} &23.3 &{23.1} &{31.3} \\
  MHFormer~\cite{li2022mhformer} ($T$=351) & CVPR'22  & 27.7 &32.1 &29.1 &28.9 & 30.0 & 33.9 &33.0 &31.2 &37.0 &39.3 & 30.0 &31.0 & 29.4 & 22.2 & 23.0 & 30.5 \\
  P-STMO~\cite{P-STMO} ($T$=243) & ECCV'22  &28.5&30.1&28.6&27.9&29.8&\textcolor{blue}{33.2}&31.3&27.8&36.0&\textcolor{blue}{37.4}&29.7&29.5&28.1&21.0&21.0&29.3\\
  MixSTE~\cite{zhang2022mixste} ($T$=243) & CVPR'22 & \textcolor{red}{21.6} & \textcolor{red}{22.0} & \textcolor{red}{20.4} & \textcolor{red}{21.0} & \textcolor{red}{20.8} & \textcolor{red}{24.3} & \textcolor{red}{24.7} & \textcolor{red}{21.9} & \textcolor{red}{26.9} & \textcolor{red}{24.9} & \textcolor{red}{21.2} & \textcolor{red}{21.5} & \textcolor{red}{20.8} & \textcolor{red}{14.7} & \textcolor{red}{15.7} & \textcolor{red}{21.6}\\
  \hline
  HSTFormer ($T$=81) & Ours & \textcolor{blue}{24.9} & \textcolor{blue}{27.4} & \textcolor{blue}{28.1} & \textcolor{blue}{25.9} & \textcolor{blue}{28.2} & 33.5 & \textcolor{blue}{28.9} & \textcolor{blue}{26.8} & \textcolor{blue}{33.4} & 38.2 & \textcolor{blue}{27.2} & \textcolor{blue}{26.7} & \textcolor{blue}{27.1} & \textcolor{blue}{20.4} & \textcolor{blue}{20.8} & \textcolor{blue}{27.8} \\
  \toprule[1pt]
  \end{tabular}
  }
  \label{table:h36m}
\end{table*}

\begin{table*}[htb]
  \normalsize
  \centering
  \caption
  {
    Quantitative comparison with the state-of-the-art methods on Human3.6M under Protocol 2 (P-MPJPE  (\textit{mm})), using detected 2D poses (CPN) as inputs. 
    \textcolor{red}{Red}: best; \textcolor{blue}{Blue}: second best. $T$ is the input length.
  } 
  \resizebox{\textwidth}{!}{
  \begin{tabular}{@{}l|c|ccccccccccccccc|c@{}}
  \toprule[1pt]
  Method (CPN) & & Dir. & Disc & Eat & Greet & Phone & Photo & Pose & Purch. & Sit & SitD. & Smoke & Wait & WalkD. & Walk & WalkT. & Avg.\\
  \midrule[0.5pt]
  ST-GCN~\cite{cai2019exploiting} ($T$=7) & ICCV'19  & 35.7 & 37.8 & 36.9 & 40.7 & 39.6 & 45.2 & 37.4 & 34.5 & 46.9 & 50.1 & 40.5 & 36.1 & 41.0 & 29.6 & 33.2 & 39.0 \\
  SGNN~\cite{zeng2021learning} ($T$=9) & ICCV'21  & 33.9 & 37.2 & 36.8 & 38.1 & 38.7 & 43.5 & 37.8 & 35.0 & 47.2 & 53.8 & 40.7 & 38.3 & 41.8 & 30.1 & 31.4 & 39.0 \\
  VPose~\cite{videopose3d} ($T$=243) & CVPR'19  & 34.1 & 36.1 & 34.4 & 37.2 & 36.4 & 42.2 & 34.4 & 33.6 & 45.0 & 52.5 & 37.4 & 33.8 & 37.8 & 25.6 & 27.3 & 36.5 \\
  Liu \emph{et al.}~\cite{liu2020attention} ($T$=243) & CVPR'20  & 32.3 & 35.2 & 33.3 & 35.8 & 35.9 & 41.5 & 33.2 & 32.7 & 44.6 & 50.9 & 37.0 & 32.4 & 37.0 & 25.2 & 27.2 & 35.6 \\
  UGCN~\cite{wang2020motion} ($T$=96) & ECCV'20  & 32.9 & 35.2 & 35.6 & 34.4 & 36.4 & 42.7 & 31.2 & 32.5 & 45.6 & 50.2 & 37.3 & 32.8 & 36.3 & 26.0 & 23.9 & 35.5 \\
  Anatomy3D~\cite{chen2021anatomy} ($T$=243) & TCSVT'21  & 32.6 &{35.1} &32.8 & 35.4 & 36.3 &{40.4} &{32.4} &32.3 &\textcolor{blue}{42.7} &{49.0} &36.8 & 32.4 & 36.0 &24.9 &26.5 &{35.0} \\
  PoseFormer~\cite{poseformer} ($T$=81) & ICCV'21  &{32.5} &34.8 & \textcolor{blue}{32.6} &34.6 &35.3 &39.5 &32.1 &32.0 &42.8 &48.5 & \textcolor{blue}{34.8} &32.4 &35.3 &24.5 &26.0 & 34.6 \\
  MHFormer~\cite{li2022mhformer} ($T$=351) & CVPR'22 & 31.5 & 34.9 & 32.8 & 33.6 & 35.3 & 39.6 & 32.0 & 32.2 & 43.5 & 48.7 & 36.4 & 32.6 & 34.3 & 23.9 & 25.1 & 34.4\\
  P-STMO~\cite{P-STMO} ($T$=243) & ECCV'22  & 31.3& 35.2& 32.9&33.9& 35.4&39.3&32.5&\textcolor{blue}{31.5}&44.6& 48.2& 36.3&32.9&34.4&\textcolor{blue}{23.8}&\textcolor{blue}{23.9}&34.4\\
  MixSTE~\cite{zhang2022mixste} ($T$=243) & CVPR'22 & \textcolor{red}{30.8}    & \textcolor{red}{33.1} & \textcolor{red}{30.3} & \textcolor{red}{31.8} & \textcolor{red}{33.1} & \textcolor{blue}{39.1}    & \textcolor{red}{31.1}    &\textcolor{red}{30.5} & \textcolor{red}{42.5}    & \textcolor{red}{44.5} & \textcolor{red}{34.0} & \textcolor{red}{30.8} & \textcolor{red}{32.7} & \textcolor{red}{22.1} & \textcolor{red}{22.9}    & \textcolor{red}{32.6} \\
  \hline
  HSTFormer ($T$=81) & Ours  & \textcolor{blue}{31.1} & \textcolor{blue}{33.7} & 33.0 & \textcolor{blue}{33.2} & \textcolor{blue}{33.6} & \textcolor{red}{38.8} & \textcolor{blue}{31.9} & \textcolor{blue}{31.5} & 43.7 & \textcolor{blue}{46.3} & 35.7 & \textcolor{blue}{31.5} & \textcolor{blue}{33.1} & 24.2 & 24.5 & \textcolor{blue}{33.7} \\
  
  \toprule[1pt]
  \end{tabular}
  }
  \label{table:h36m_p2}
\end{table*}

\section{Experiments}
\subsection{Datasets and Evaluation Metrics}
The proposed method is evaluated on three benchmark datasets, \textit{i.e.}, Human3.6M \cite{ionescu2013human3}, MPI-INF-3DHP \cite{3dhp}, and HumanEva \cite{sigal2010humaneva}.

\noindent\textbf{Human3.6M.} It is the most representative benchmark dataset for 3D human pose estimation. It contains over 3.6 million images and their corresponding 3D human pose annotations. 11 subjects perform 15 daily activities in an indoor environment, such as walking, sitting, smoking, \textit{etc.} High-resolution videos of each subject are recorded from four different views. Following the same policy of others \cite{poseformer,li2022mhformer,zhang2022mixste,P-STMO}, 5 subjects (S1, S5, S6, S7, S8) are used for training and 2 subjects (S9, S11) are used for testing.

\noindent\textbf{MPI-INF-3DHP.} It is also a widely-used large-scale and more challenging dataset containing both indoor and complex outdoor scenes for 3D human pose estimation. It has 1.3 million frames with more diverse motions than Human3.6M. The training set has 8 subjects performing 8 activities and the test set has 7 subjects. The same as SOTAs \cite{poseformer,li2022mhformer,zhang2022mixste,P-STMO}, we train our method using the training set and evaluate it using the valid frames in the test set.

\noindent\textbf{HumanEva.} It is a  smaller but challenging dataset compared to the above two datasets. The HumanEva consists of about 29,000 frames from 7 calibrated video sequences. Given the small volume, it becomes challenging in converging the training model to achieve good performance. Following the same setting to \cite{poseformer,zhang2022mixste}, two actions including Walking and jogging in subjects S1, S2, and S3 are evaluated.

\noindent\textbf{Evaluation Metrics.} 
Following \cite{poseformer,li2022mhformer,zhang2022mixste,P-STMO}, we use the same metrics for performance evaluation. For Human3.6M, two evaluation protocols are adopted to calculate the quantitative results. Protocol 1 refers to MPJPE which is the mean Euclidean distance between the predictions and ground truths in millimeters  (\textit{mm}). Protocol 2 refers to P-MPJPE which is the MPJPE between aligned 3D pose predictions and ground truths. For MPI-INF-3DHP, the area under the curve (AUC), percentage of correct keypoints (PCK), and MPJPE are used as
evaluation metrics. For HumanEva, MPJPE is used for performance evaluation.

\begin{figure}[t]
	\centering
	\includegraphics[width=0.47\textwidth]{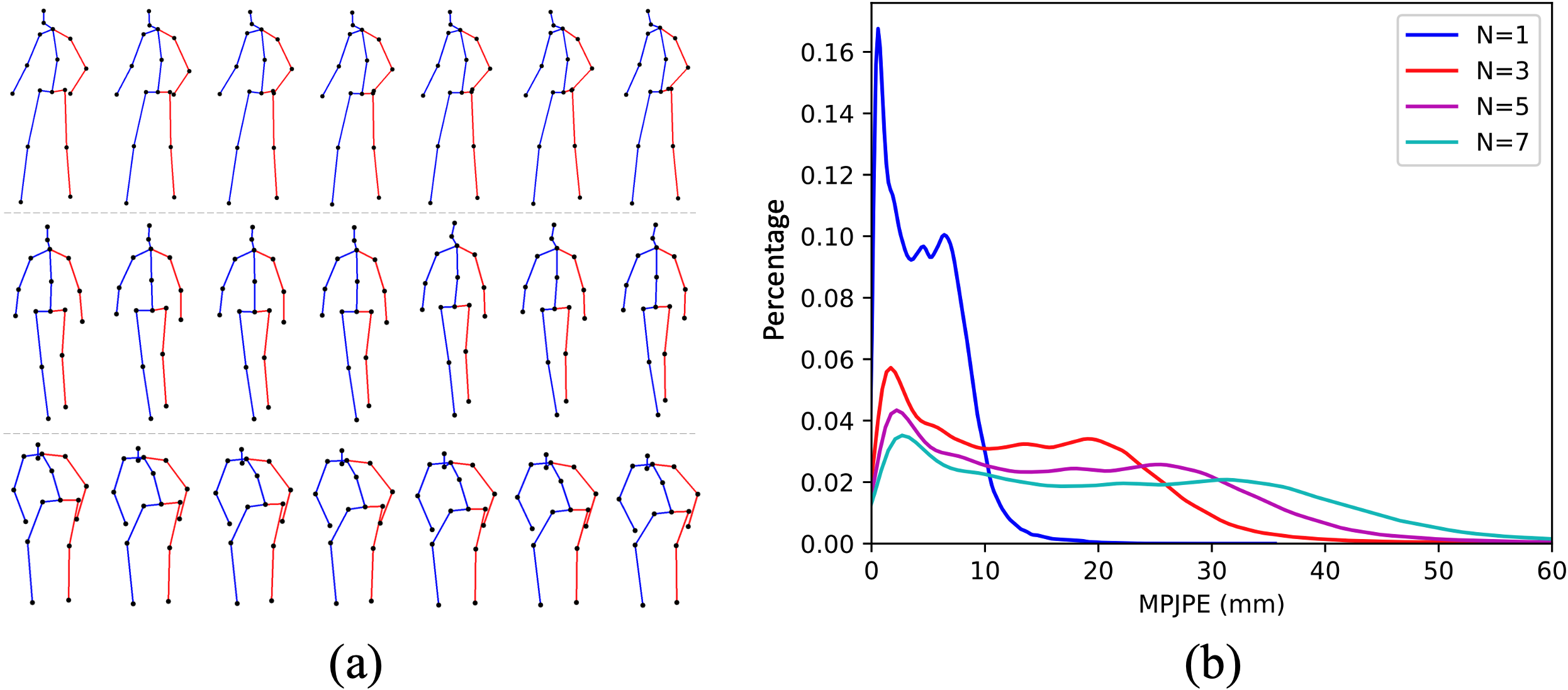}
	\caption{ (a) Visual examples of 2D pose sequence with consecutive frames. (b) MPJPE statistic results obtained by computing the MPJPE between the 2D pose ground truth of adjacent frames with a sample interval $N$ on the Human3.6M dataset. }
	\label{fig.mpjpe}
\end{figure}

\subsection{Implementation Details}
HSTFormer is implemented in PyTorch. We train the HSTFormer from scratch for 100 epochs on 4 Nvidia Tesla V100 GPUs using Adam optimizer with an initial learning rate of 0.001 and decaying it to 80\% for every 20 epochs. The batchsize is 128 for each GPU. The poses are flipped horizontally to perform data augmentation for both training and testing, following \cite{pavllo2020modeling, poseformer}.

We observe that the poses from adjacent frames usually contain redundant information since few changes happen between them if the video FPS is large (25 or 30). Fig. \ref{fig.mpjpe} (a) shows examples of three actions with seven consecutive frames, visually demonstrating this observation. Hence, to increase the diversity of the input poses, one way is to collect a longer pose sequence. But this will increase the computation burden. Another way is to collect a pose sequence with a fixed number of frames $T$ and sample the frames with an interval $N$ (\textit{i.e.}, $f_{i+1}=f_i + N$, where $i=1,...,T-1$ is the pose index in the collected pose sequence and $f_i$ is the frame index in the video), meaning that the pose sequence will cover more temporal information with fixed computation cost when giving a larger $N$. Therefore, we adopt the second pose sequence collection way in this work. The distribution of MPJPEs computed between adjacent frames with a sample interval of 1, 3, 5, and 7 are shown in Fig. \ref{fig.mpjpe} (b). We can see that a larger sample interval produces a more diverse pose sequence of a bigger MPJPE value. 

\subsection{Comparison with State-of-the-art Models}
\noindent\textbf{Results on Human3.6M.}
The proposed HSTFormer is compared with existing SOTA approaches on the Human3.6M dataset. We trained HSTFormer with the following setting: $T=81$, $N=7$, and $N_{en}=6$. As done by others, HSTFormer also takes the sequences of 2D pose predictions detected by the popular cascaded pyramid network (CPN)  \cite{chen2018cascaded} and ground truths (GT) as inputs for training. Table \ref{table:h36m} and Table \ref{table:h36m_p2} show their best reported results of different methods in terms of Protocol 1 (MPJPE) using both CPN and GT inputs and Protocol 2 (P-MPJPE) using CPN input, respectively. It can be seen that our method achieves the second best performance for all overall comparisons even its input sequence length is $T=81$. Additionally, our average MPJPE (27.8 \textit{mm}) outperforms the third best result (29.3 \textit{mm}) by 5.1\% (1.5 \textit{mm}) when taking GT poses as input. These demonstrate the competitiveness and effectiveness of the proposed method on the constrained and simple Human3.6M dataset. It is also worth pointing out that our result is about 6.2 \textit{mm} in average worse than the best result MixSTE. However, we would argue that our approach is more generalized and consistent, which is proven in the following sections.

\noindent\textbf{Results on MPI-INF-3DHP.}
To verify the generalization ability of different methods on a more challenging large-scale dataset, we evalaute the proposed method on MPI-INF-3DHP by taking GT poses as input and compare its results with SOTA approaches. We observe that P-STMO \cite{P-STMO} and others (\textit{e.g.}, PoseFormer \cite{poseformer}, MHFormer \cite{li2022mhformer}, and MixSTE \cite{zhang2022mixste}) use different 3D pose annotations during training. For fair comparison, we train our model using both data with the settings $T=27/81$, $N=7$, and $N_{en}=6$. Table \ref{tab:3dhp} lists our results and the best reported results of others on the MPI-INF-3DHP testset. It can be seen that our method HSTFormer achieves the best performance in three evaluation metrics for all comparisons when $T=81$ and even the second best when $T=27$. Specifically, HSTFormer outperforms MixSTE \cite{zhang2022mixste} (the best method on Human3.6M) by a large margin, \textit{e.g.}, decreasing MPJPE by 24.6\% (from 54.9 \textit{mm} to 41.4 \textit{mm}). This demonstrates that our method is more powerful to handle the challenging and complex outdoor scenes than others. 

We further examine the performance comparison of same method evaluated on different datasets and have the following observation. Comparing to the results in the bottom part of Table \ref{table:h36m} and Table \ref{tab:3dhp}, we notice that the performance of the SOTA methods drops much more than ours when evaluating on Human3.6M and MPI-INF-3DHP. For example, the average MPJPE is increased by 27.5 \textit{mm} (90.2\%) of MHFormer \cite{li2022mhformer} and 23.3 \textit{mm} (107.9\%) of MixSTE \cite{zhang2022mixste} vs 13.6 \textit{mm} (48.9\%) of ours. There are two conclusions drawn from the above observation. First, the proposed HSTFormer has a higher generalization ability than others on challenging and complex scenes. Second, the current SOTAs are prone to overfit on a relative more saturated Human3.6M dataset.

\begin{figure}[t]
	\centering
	\includegraphics[width=0.48\textwidth]{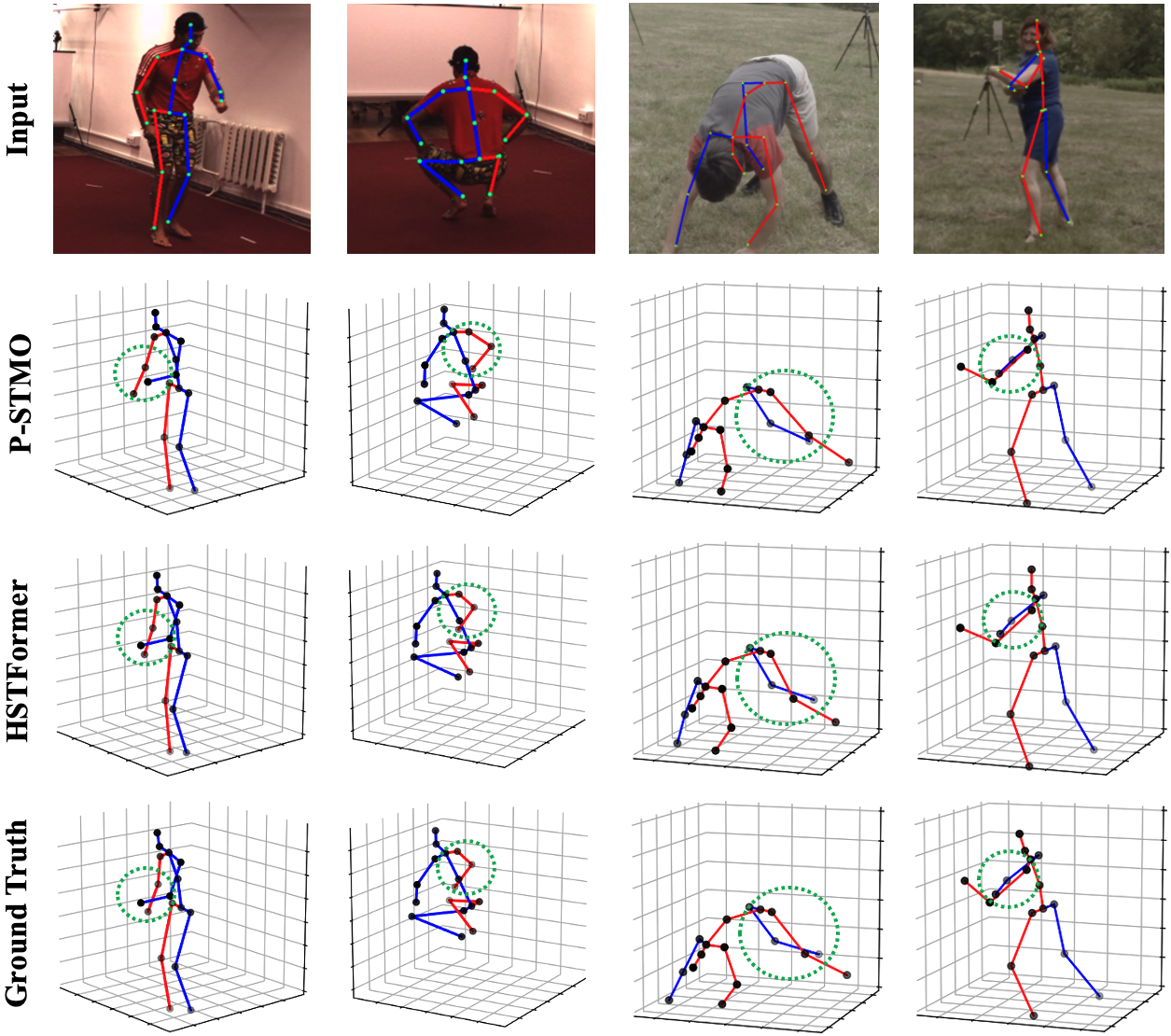}
	\caption{Qualitative comparisons between HSTFormer and a state-of-the-art approach P-STMO \cite{P-STMO} on Human3.6M (the left two columns) and  MPI-INF-3DHP (the right two columns). The green circles highlight the differences between their 3D pose estimation results and the ground truths.}
	\label{fig.visulization}
\end{figure}

\begin{table}[tp]
		\centering
		\caption{Quantitative comparison with the state-of-the-art methods on MPI-INF-3DHP under three metrics. $\uparrow$ indicates the higher, the better. $\downarrow$ indicates the lower, the better. $^\star$ means the model is trained using the same data as P-STMO \cite{P-STMO}. \textcolor{red}{Red}: best; \textcolor{blue}{Blue}: second best. $T$ is the input length.}
		\resizebox{220pt}{!}{
			\begin{tabular}{l|c|ccc}
				\toprule
				Method                               &                & PCK$\uparrow$          & AUC$\uparrow$          & MPJPE$\downarrow$        \\
				\midrule
				Lin \etal~\cite{lin2019trajectory} ($T$=25) 	 & BMVC'19       & 83.6          & 51.4          & 79.8          \\
				Chen \etal~\cite{anatomypose} ($T$=243)           & TCSVT'21      & 87.8          & 53.8          & 79.1          \\
				PoseFormer~\cite{poseformer} ($T$=9)                  & ICCV'21       & 88.6          & 56.4          & 77.1 \\
				Wang \etal~\cite{motionguidepose} ($T$=96)        & ECCV'20       & 86.9          & 62.1          & 68.1          \\
				MHFormer~\cite{li2022mhformer} ($T$=9) & CVPR'22 & 93.8 & 63.3 & 58.0\\
				MixSTE~\cite{zhang2022mixste} ($T$=27)              & CVPR'22      & 94.4 & 66.5 & 54.9	\\
				\hline
				HSTFormer ($T$=27) & Ours & \textcolor{blue}{96.6} & \textcolor{blue}{70.5} & \textcolor{blue}{44.5}	\\ 
				HSTFormer ($T$=81) & Ours & \textcolor{red}{97.3} & \textcolor{red}{71.5} & \textcolor{red}{41.4}	\\ 
				\hline
				\hline
				P-STMO~\cite{P-STMO}$^\star$ ($T$=81) & ECCV'22 &97.9&75.8&32.2\\
				\hline
				HSTFormer$^\star$ ($T$=27) & Ours & \textcolor{red}{98.0} & \textcolor{blue}{78.4} & \textcolor{blue}{28.6} \\ 
				HSTFormer$^\star$ ($T$=81) & Ours & \textcolor{red}{98.0} & \textcolor{red}{78.6} & \textcolor{red}{28.3}	\\ 
				\bottomrule
			\end{tabular}
		}
		\label{tab:3dhp}
	\end{table}

\begin{table}[tp]
	\centering
	\caption{Quantitative comparison with the state-of-the-art methods on HumanEva under MPJPE  (\textit{mm}). $^\dagger$ means the model is retrained using its default setting and the same data as others. \textcolor{red}{Red}: best; \textcolor{blue}{Blue}: second best. $T$ is the input length.}
	\resizebox{0.48\textwidth}{!}{
		\begin{tabular}{l|ccc|ccc|c}
			\toprule
			\multirow{2}{*}{Method} & \multicolumn{3}{c|}{Walk}                      & \multicolumn{3}{c|}{Jog}                       &\multirow{2}{*}{Avg.}           \\ \cline{2-7}
			& S1 & S2 & S3 & S1 & S2 & S3 & \\
			\midrule
			MixSTE~\cite{zhang2022mixste} ($T$=43)           & 20.3          & 22.4          & 34.8          & 27.3          & 32.1          & 34.3          & 28.5          \\
			MHFormer~\cite{li2022mhformer}$^\dagger$ ($T$=43)           &   20.6        &   14.6        &  \textcolor{red}{32.7}         &   34.1        &   20.6        &  23.8         &    24.4       \\
            PoseFormer~\cite{poseformer} ($T$=43)           &   \textcolor{blue}{16.3}        &  \textcolor{red}{11.0}         &  45.7         &   \textcolor{red}{25.0}        &  \textcolor{red}{15.2}         &  \textcolor{red}{15.1}         & 21.6          \\
\hline
            HSTFormer ($T$=9)           & 16.7          &  14.3         & 34.2          &  25.9         & 18.1          &  19.4         & \textcolor{blue}{21.4}          \\
            HSTFormer ($T$=43)           & \textcolor{red}{15.4}          &  \textcolor{blue}{12.0}         &  \textcolor{blue}{33.2}         &  \textcolor{blue}{25.5}         & \textcolor{blue}{16.5}          &  \textcolor{blue}{18.4}         & \textcolor{red}{20.2}          \\
			\bottomrule
		\end{tabular}
	}
	\label{tab:humaneva_p1}
\end{table}

\noindent\textbf{Results on HumanEva.} 
To further explore the generalization ability of different methods on a smaller dataset, we evaluate our method HSTFormer on the HumanEva dataset. The challenge of a small dataset such as HumanEva is due to its short video lengths. This particularly requires more robust spatial and temporal correlations of joints to avoid the overfitting during training.
We configure our model with the following settings: $T=9/43$, $N=1$, and $N_{en}=6$. All compared methods and ours are trained from scratch on HumanEva. Table \ref{tab:humaneva_p1} reports the MPJPE results of different methods. It can be seen that HSTFormer achieves the best performance on both $T=9$ and $T=43$ compared to others. Specifically, it outperforms PoseFormer \cite{poseformer} by 6.5\% (1.4 \textit{mm}), MHFormer \cite{li2022mhformer} by 17.2\% (4.2 \textit{mm}), and MixSTE \cite{zhang2022mixste} by 29.1\% (8.3 \textit{mm}) in average MPJPE. This demonstrates that HSTFormer is more general or suitable than the others on small datasets due to comprehensively capturing joints' multi-level spatial and temporal correlations.

\noindent\textbf{Qualitative Results.}
As shown in Fig. \ref{fig.visulization}, we also visualize some 3D pose estimation results produced by HSTFormer and P-STMO \cite{P-STMO} on both Human3.6M and MPI-INF-3DHP datasets. This visually demonstrates that our model can estimate accurate 3D poses in both constrained indoor and complex outdoor scenes/activities, especially on the parts of limbs that may move fast or be occluded. 
\subsection{Ablation Study}
To verify the effect of different modules and hyper-parameters in HSTFormer, extensive ablation experiments are conducted on Human3.6M with GT poses as input and MPJPE (\textit{mm}) as evaluation metric. 

\noindent\textbf{Effect of Architecture Modules.}
We first study the module choices of HSTFormer by configuring them with different combinations. The results are reported in Table \ref{table:component}. It can be seen that the model makes steady and accumulated improvement in MPJPE (from 41.6 \textit{mm} to 30.6 \textit{mm}) as more modules are hierarchically built on with temporal emphasis and fusion. The joint-wise correlations information is effectively modeled as more modules are included and aggregated. Notably, (i) the model gets the largest performance improvement (from 39.8 \textit{mm} to 35.1 \textit{mm}) when adding BTTE. This demonstrates that the explicitly modeling of the body parts' local motions is crucial. (ii) The fusion module plays an essential role in improving the MPJPE performance from 33.4 \textit{mm} to 30.6 \textit{mm}. 

Moreover, we investigate two module aggregation strategies: consecutively aggregating the encoders (i) from local to global and (ii) from global to local. Table \ref{tab:ablation_strategy} reports their results. It is apparent that aggregating the modules from local to global is much better than the other. This rigorously proves that hierarchical modeling is effective, following the obvious kinesiological analysis (joints to body-parts, body-parts to poses). The correct hierarchical order helps to pass the crucial information for better performance, while the incorrect order jeopardizes the result.

\begin{table}[t]
  \centering
  \caption
  {
    Ablation study on different encoder combinations.
  }
  \resizebox{0.4\textwidth}{!}{
  \setlength{\tabcolsep}{2.10mm}
  \begin{tabular}{cccccc|c}
  \toprule  [1pt]
  Model & STE & JTTE & BTTE & PTTE & Fusion &MPJPE \\
  \midrule [0.5pt]
  1 & \textcolor{red}{\cmark} &\xmark &\xmark &\xmark &\xmark &41.6 \\
  2 & \textcolor{red}{\cmark} &\textcolor{red}{\cmark} &\xmark &\xmark &\xmark &39.8 \\
  3 & \textcolor{red}{\cmark} &\textcolor{red}{\cmark} &\textcolor{red}{\cmark} &\xmark &\xmark &35.1 \\
  4 & \textcolor{red}{\cmark} &\textcolor{red}{\cmark} &\textcolor{red}{\cmark} &\textcolor{red}{\cmark} &\xmark &33.4 \\
  5 & \textcolor{red}{\cmark} &\textcolor{red}{\cmark} &\textcolor{red}{\cmark} &\xmark &\textcolor{red}{\cmark} &32.2 \\
  6 & \textcolor{red}{\cmark} &\textcolor{red}{\cmark} &\textcolor{red}{\cmark} &\textcolor{red}{\cmark} &\textcolor{red}{\cmark} &30.6 \\

  \toprule [1pt]
  \end{tabular}
  }
  \label{table:component}
\end{table}

\begin{table}[t]
	\centering
	\caption{Ablation study for different component aggregation strategies.}
	\resizebox{0.38\textwidth}{!}{
		\begin{tabular}{c|c}
			\toprule
			Aggregation strategy & MPJPE \\
			\midrule
		    (i) STE $\Rightarrow$ JTTE $\Rightarrow$ BTTE $\Rightarrow$ PTTE $\Rightarrow$ Fusion    &   30.6    \\
			(ii) PTTE $\Rightarrow$ BTTE $\Rightarrow$ JTTE $\Rightarrow$ STE $\Rightarrow$ Fusion    &   37.4    \\
			\bottomrule
		\end{tabular}
	}
	\label{tab:ablation_strategy}
\end{table}

\begin{table}[t]
	\centering
	\caption{Ablation study for hyper-parameter setting in the number of TE layers ($N_{en}$) and input length ($T$).}
	\resizebox{0.34\textwidth}{!}{
		\begin{tabular}{cc|c|c|c}
			\toprule
			$N_{en}$ & $T$ & Params (M) & FLOPs (G) & MPJPE \\
			\midrule
			4      & 81    & 15.25  & 2.64 &  32.8    \\
			6      & 81    & 22.81  & 3.94 &  30.6    \\
			8      & 81    & 30.38  & 5.24 &  30.4    \\
			\hline
			6      & 9     & 22.73  & 0.45 &  35.5    \\
			6      & 27    & 22.75  & 1.31 &  34.4    \\
			6      & 43    & 22.77  & 2.10 &  31.7    \\
			\bottomrule
		\end{tabular}
	}
	\label{tab:ablation_params}
\end{table}

\begin{table}[t]
	\centering
	\caption{Ablation study for hyper-parameter setting in the input length ($T$) and interval ($N$).}
	\setlength\tabcolsep{2pt}
	\resizebox{0.48\textwidth}{!}{
		\begin{tabular}{c|cccc|cccc|cccc|cccc}
			\toprule
			$T$ & \multicolumn{4}{c|}{9} & \multicolumn{4}{c|}{27}   & \multicolumn{4}{c|}{43} & \multicolumn{4}{c}{81} \\ \midrule 
			$N$ & 1 & 3 & 5 & 7 & 1 & 3 & 5 & 7 & 1 & 3 & 5 & 7 & 1 & 3 & 5 & 7 \\
\hline
            MPJPE & 35.5 & 34.4 & 33.4 & 32.5 & 34.4 & 32.3 & 31.2 & 30.4 & 31.7 & 30.2 & 29.3 & 28.7 & 30.6 & 28.9 & 28.3 & 27.8 \\
			\bottomrule
		\end{tabular}
	}
	\label{tab:interval}
\end{table}

\noindent\textbf{Effect of Architecture Hyper-Parameters.}
Table \ref{tab:ablation_params} reports the results of different settings of the hyper-parameter $N_{en}$ (the number of TE layers) and $T$ (the input length). It can be seen that although $N_{en}=8$ and $T=81$ can get a little better performance than $N_{en}=6$ and $T=81$, we select the latter one instead of the former one as the final setting because the former one introduces more parameters (30.38M vs 22.81M) and requires more computation (5.24G vs 3.94G).
Table \ref{tab:interval} reports the results of different settings of the hyper-parameter $T$ (the input length) and $N$ (the sample interval). It is obvious that the model performance of MPJPE is improved as the frame sampling interval becomes larger while fixing the input length. We also validate the setting of $T=81$ and $N=9$ in our model and find that the performance of MPJPE is not better than the setting of $T=81$ and $N=7$. Therefore, we select $N=7$ in this work. This demonstrates that we can improve the performance but without introducing extra computation cost and model complexity by fixing the input length and adopting a large sample interval, so as to obtain a relatively large temporal receptive field.

\section{Conclusions}
In this paper, we present HSTFormer, a novel framework based on hierachical spatial-temporal transformers for 3D human pose estimation from 2D pose sequence. HSTFormer processes pose sequence in a hierarchical paradigm, from joints to body-parts, and eventually to the entire pose. Four transformer encoders are concatenated following the kinesiological orders: spatial transformer encoder, joint temporal transformer encoder, body-part temporal transformer encoder, and pose temporal transformer encoder.  In addition, a fusion model gathers all layers is used in predict the 3D poses. Extensive experiments and detailed ablation studies are conducted. A superiror and consistent performance is demonstrated with the effectiveness of the hierarchical design. Our method outperforms state-of-the-art approaches by a large margin on the challenging and MPI-INF-3DHP dataset with complex outdoor scenes. 

\appendix
\section{Comparison of Computational Complexity}
Table \ref{tab:detailed_comp} lists the number of parameters, floating-point operations (FLOPs) at inference, and MPJPE evaluated on Human3.6M \cite{ionescu2013human3} and MPI-INF-3DHP\cite{3dhp} with GT poses as input to compare the computational complexity of different methods. The parameter number and FLOPs are calculated using the functions (\texttt{parameter\_count\_table()} and \texttt{FlopCountAnalysis()}) from the fvcore package\footnote{\url{https://github.com/facebookresearch/fvcore}}. The best MPJPEs reported in their papers are listed in the last column for reference.

From Table \ref{tab:detailed_comp}, there are two observations. 
First, the HSTFormer achieves the best performance on both datasets compared to the others except MixSTE on Human3.6M. 
Three temporal-attention modules are added in our method. 
This leads to that the proposed HSTFormer has more parameters than PoseFormer \cite{poseformer} and P-STMO \cite{P-STMO} but is still less than MHFormer \cite{li2022mhformer} and MixSTE \cite{zhang2022mixste}. 
This demonstrates that the effectiveness of the proposed HSTFormer is not only because of introducing more parameters, but more importantly because of its capability of learning the multi-level spatial-temporal correlations of joints comprehensively and structurally. 

Second, HSTFormer outperforms MixSTE on MPI-INF-3DHP by a large margin, \textit{e.g.}, decreasing MPJPE by 24.6\% (from 54.9 \textit{mm} to 41.4 \textit{mm}), but gets worse results than MixSTE on Human3.6M. The possible reason is that MixSTE is prone to overfit on the relative saturated Human3.6M dataset and our method has a high generalization ability on the challenging MPI-INF-3DHP dataset, which has been discussed in the manuscript.

\begin{table}[t]
	\centering
	\caption{Comparisons of different methods in terms of parameters, FLOPs, and MPJPE. The evaluation is performed on Human3.6M/MPI-INF-3DHP with GT poses as input. $^\star$ means the result is obtained by a model trained using the same data as P-STMO \cite{P-STMO} on MPI-INF-3DHP. $T$ is the input length.}
	\resizebox{0.48\textwidth}{!}{
		\begin{tabular}{c|c|c|c}
			\toprule
			Method & Params (M) & FLOPs (G)  & MPJPE \\
			\midrule
			PoseFormer~\cite{poseformer} ($T$=81/9)    & 9.6/9.6  & 0.8/0.1  & 31.3/77.1   \\
			MHFormer~\cite{li2022mhformer} ($T$=351/9)    & 31.5/19.1  & 8.2/0.2   & 30.5/58.0  \\
			MixSTE~\cite{zhang2022mixste} ($T$=243/27)    & 33.8/33.7 & 147.9/15.6  & 21.6/54.9    \\
			\hline
			HSTFormer ($T$=81/81)    &  22.72/22.72  & 2.12/2.12  & 27.8/41.4   \\
			\hline
			\hline
			P-STMO~\cite{P-STMO} ($T$=243/81)    & 8.9/8.0  & 1.4/0.5  & 29.3/32.2$^\star$   \\
			\hline
			HSTFormer ($T$=81/81)    &  22.72/22.72  & 2.12/2.12  & 27.8/28.3$^\star$   \\
			\bottomrule
		\end{tabular}
	}
	\label{tab:detailed_comp}
\end{table}

\section{More Qualitative Results on the Human3.6M and MPI-INF-3DHP Datasets}
\begin{figure*}[ht]
	\centering
	\includegraphics[width=0.999\textwidth]{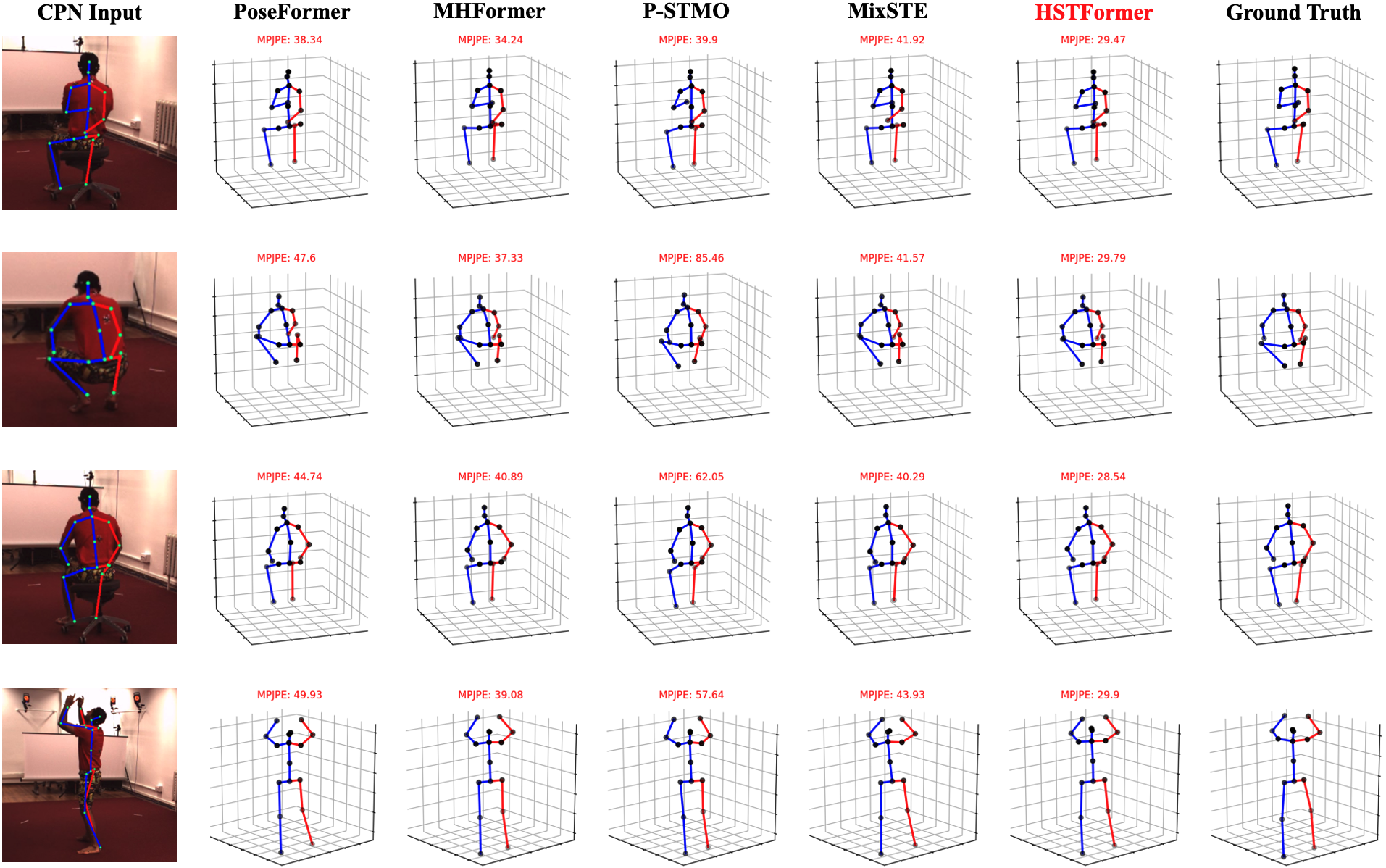}
	\caption{Qualitative comparison among the proposed method HSTFormer and the state-of-the-art approaches (PoseFormer, MHFormer, MixSTE, and P-STMO) on Human3.6M.}
	\label{fig.s1}
\end{figure*}

The existing approaches PoseFormer \cite{poseformer}, MHFormer \cite{li2022mhformer}, MixSTE \cite{zhang2022mixste}, and P-STMO \cite{P-STMO} release their trained models on Human3.6M with CPN poses as input. Thus, we provide various qualitative comparisons of the propsoed HSTFormer and the aforementioned SOTA approaches on Human3.6M. This is shown in Figure \ref{fig.s1}. 

\begin{figure*}[ht]
	\centering
	\includegraphics[width=0.999\textwidth]{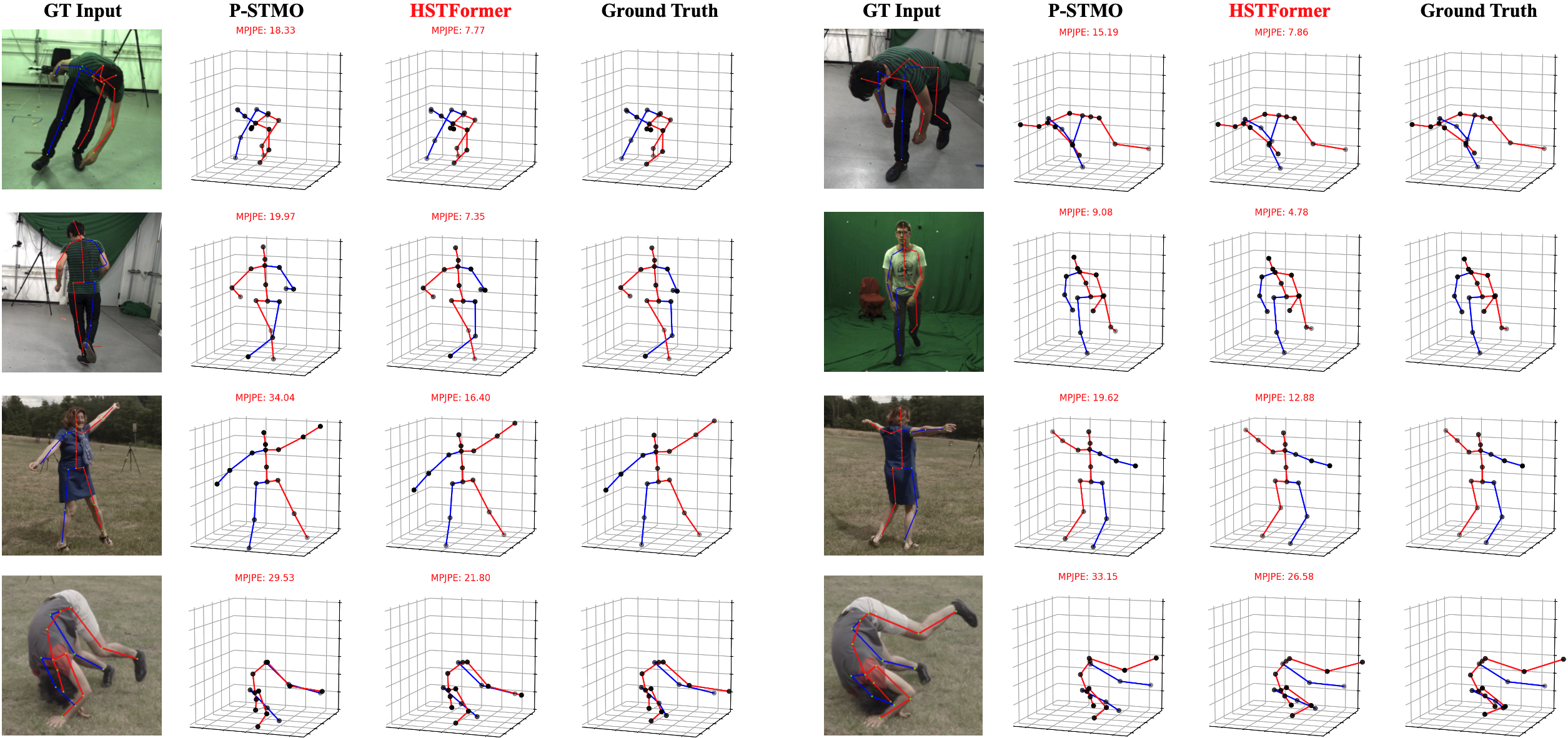}
	\caption{Qualitative comparison among the proposed method HSTFormer and the state-of-the-art approach P-STMO on MPI-INF-3DHP.}
	\label{fig.s2}
\end{figure*}

In addition, we take analysis on the the MPI-INF-3DHP dataset as well. Since only the P-STMO \cite{P-STMO} provides the trained model on MPI-INF-3DHP with GT poses as input. Thus, we compare the qualitative results of P-STMO and ours on MPI-INF-3DHP, as shown in Figure \ref{fig.s2}.

From Figure \ref{fig.s1} and Figure \ref{fig.s2}, it can be seen that our method is able to produce accurate 3D poses in both constrained indoor and challenging outdoor scenes/activities, especially on the parts of limbs that may move in various speed, as well as be occluded.

\subsection{Video Demos}

\begin{figure*}[!t]
	\centering
	\includegraphics[width=0.98\textwidth]{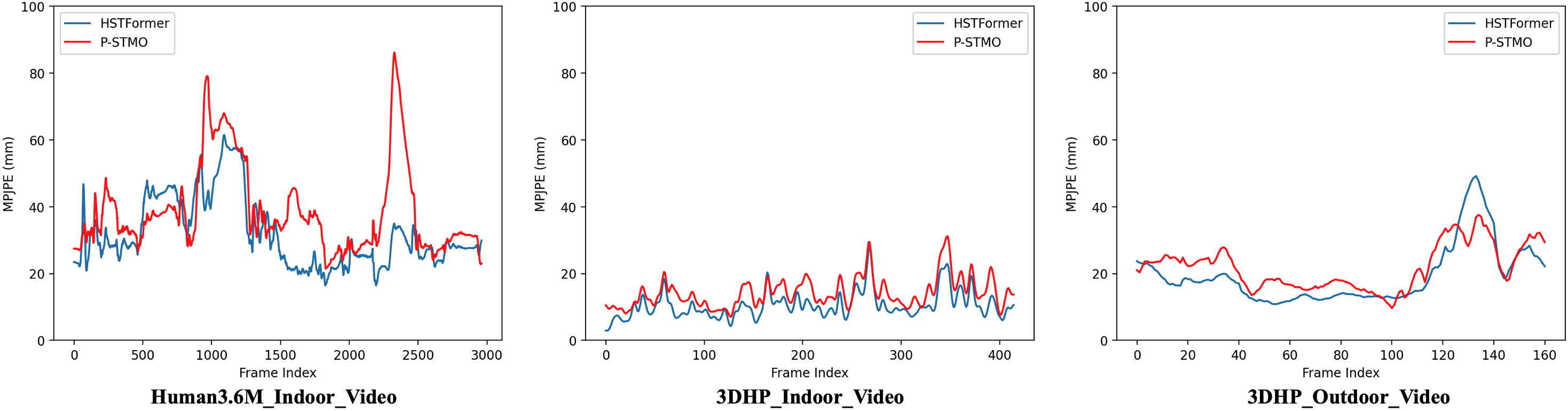}
	\caption{This is associated with the video filenames ``Human3.6M\_Indoor\_Video.mp4",  ``3DHP\_Indoor\_Video.mp4", and ``3DHP\_Outdoor\_Video.mp4", respectively. 
	The figure shows the frame-wise MPJPE comparisons between the proposed method HSTFormer and the state-of-the-art approach P-STMO on three videos.  }
	\label{fig.mpjpe}
\end{figure*}

We attached three videos\footnote{\url{https://drive.google.com/drive/folders/1nffpLTzBrvmZp0MBH6WEoLu_YVtxWwpw?usp=share_link}} with following filenames:
\begin{itemize}
  \item[--] ``Human3.6M\_Indoor\_Video.mp4" 
  \item[--] ``3DHP\_Indoor\_Video.mp4"
  \item[--] ``3DHP\_Outdoor\_Video.mp4"
\end{itemize}
with side-by-side comparison of the proposed HSTFormer and P-STMO. In order to further demonstrate the superior performance of the proposed HSTFormer, we further plot frame-wise average MPJPE across all estimated joints which re-depicted in Fig. \ref{fig.mpjpe}. These results are associated with the video demo filenames in the following, respectively.
\begin{itemize}
  \item[--] ``Human3.6M\_Indoor\_Video.mp4"
  \item[--] ``3DHP\_Indoor\_Video.mp4"
  \item[--] ``3DHP\_Outdoor\_Video.mp4"
\end{itemize}
Our model (blue line) achieves better performance than P-STMO. Specifically, our model reconstruct 3D pose with lower MPJPE in most frames. It also shows that our model is more stable and robust with better consistency.

\begin{figure*}[!t]
	\centering
	\includegraphics[width=0.999\textwidth]{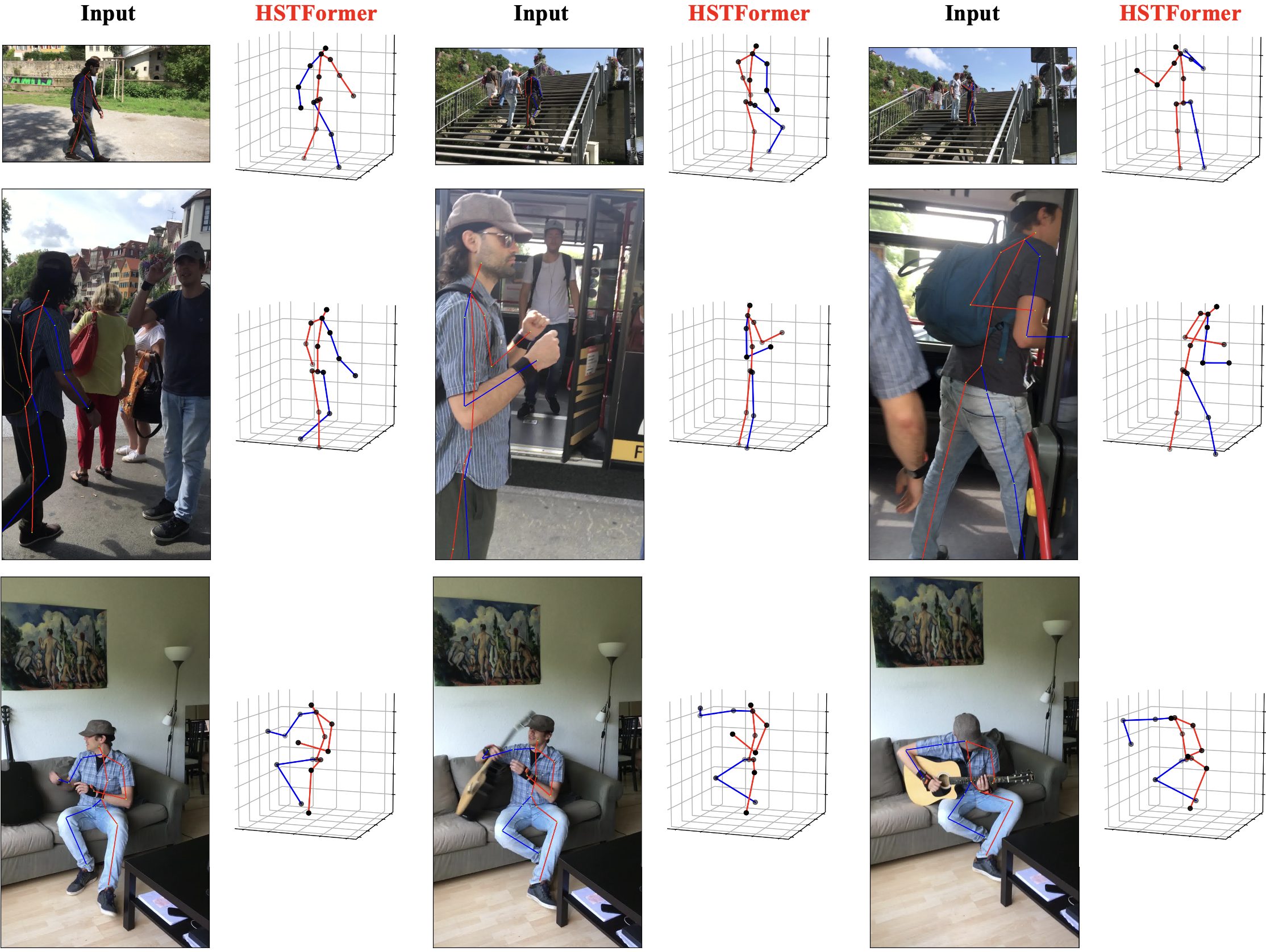}
	\caption{Qualitative results of the proposed method HSTFormer on ``in-the-wild'' videos from 3DPW.}
	\label{fig.s3}
\end{figure*}

\section{Qualitative Results on ``in-the-wild'' Videos}
To verify the effectiveness and generalization ability of the proposed method on ``in-the-wild'' videos that often contain unseen poses/activities in the training set, we run our method HSTFormer on the 3DPW \cite{von2018recovering} dataset. Here, HSTFormer is trained using the MPI-INF-3DHP training set. 
Figure \ref{fig.s3} shows some visual examples of 3D pose estimation results produced by HSTFormer. Additionally, two video demos are also included in the supplementary material, named ``3DPW\_Video\_1.mp4" and ``3DPW\_Video\_2.mp4". From them, we can see that HSTFormer can generalize well to ``in-the-wild'' videos, even for the unseen poses/activities in the training set.

{\small
\bibliographystyle{ieee_fullname}
\bibliography{egbib}
}

\end{document}